\DocumentMetadata{
	lang		= eng-US, 	
	pdfstandard	= A-3u,		
	pdfversion	= 1.7, 		
}

\documentclass[colorlinks,upint,subscriptcorrection,varvw,hyphenate,balance,greek,russian,vietnamese,german]{asmeconf} 

\hypersetup{%
	pdfauthor={Tian-Ao Ren},									  
	pdftitle={Multimodal Sensing for Robot-Assisted Sub-Tissue Feature Detection in Physiotherapy Palpation},                  
	pdfkeywords={Tactile sensing, force sensing, multimodal sensing, medical palpation, tissue manipulation},
	pdfsubject = {Robotic multimodal sensing methods for detecting sub-tissue features in physiotherapy palpation},			  
}

\allowdisplaybreaks 

\begin{document}


\ConfName{Proceedings of the 2026 Design of Medical Devices Conference}
\ConfAcronym{DMD2026}
\ConfDate{April 20--22, 2026} 
\ConfCity{Minneapolis, MN}
\PaperNo{DMD2026-1039}

%

\title{Multimodal Sensing for Robot-Assisted Sub-Tissue\\
Feature Detection in Physiotherapy Palpation}
 
%
%
%

\SetAuthors{%
    Tian-Ao Ren\affil{1}\CorrespondingAuthor{},
	Jorge Garcia\affil{1}, 
    Seongheon Hong\affil{1}, \\
    Jared Grinberg\affil{2},
	Hojung Choi\affil{1}, 
    Julia Di\affil{1},
    Hao Li\affil{1},
    Dmitry Grinberg\affil{2}, Mark R. Cutkosky\affil{1}\CorrespondingAuthor{tianao@stanford.edu, cutkosky@stanford.edu}
	}

\SetAffiliation{1}{Stanford University, Stanford, CA}
\SetAffiliation{2}{Symbiokinetics Inc, Palo Alto, CA}



\maketitle

\versionfootnote{Documentation for \texttt{asmeconf.cls}: Version~\versionno, \today.}


\keywords{Tactile sensing, force sensing, multimodal sensing, medical palpation, tissue manipulation}


\begin{abstract}
Robotic palpation relies on force sensing, but force signals in soft-tissue environments are variable and cannot reliably reveal subtle subsurface features. We present a compact multimodal sensor that integrates high-resolution vision-based tactile imaging with a 6-axis force–torque sensor. 
In experiments on silicone phantoms with diverse subsurface tendon geometries, force signals alone frequently produce ambiguous responses, while tactile images reveal clear structural differences in presence, diameter, depth, crossings, and multiplicity. Yet accurate force tracking remains essential for maintaining safe, consistent contact during physiotherapeutic interaction. Preliminary results show that combining tactile and force modalities enables robust subsurface feature detection and controlled robotic palpation.
\end{abstract}



\section{Introduction}
A foundational diagnostic skill, palpation enables clinical practitioners to assess tissue stiffness, localize tendons, and identify abnormalities beneath soft tissue \cite{johansson1984roles, myburgh2008systematic}. Clinicians naturally integrate applied force with tactile perception, adjusting pressure while interpreting local deformation cues. Replicating this capability in robotic systems remains challenging. Traditional robotic palpation methods rely on force sensing (e.g. \cite{zhang2020piezoelectric, gwilliam2009effects}),
but force signals can be difficult to interpret in soft-tissue environments.

In the present context, we are particularly interested in physiotherapy applications (e.g.~acupressure \cite{harada2025muscle, xu2018human, konstantinova2016autonomous}) which combine palpation with large, targeted applications of 
pressure, necessitating both accurate force control and robust hardware that can apply and withstand forces of 100\,N routinely.

The ambiguity arising from force/torque contact data is 
evident even in controlled settings: as shown later, 
sliding across a silicone phantom while pressing firmly can produce substantially different force profiles depending on the local geometry as well as whether there are subsurface features. Variations in force profiles can also arise from fine-scale sub-surface features, especially when geometry differences involve small height changes, lateral variations, crossings, or tendon multiplicity \cite{guo2019compensating}.

\begin{figure}
\centering\includegraphics[width=0.9\linewidth]{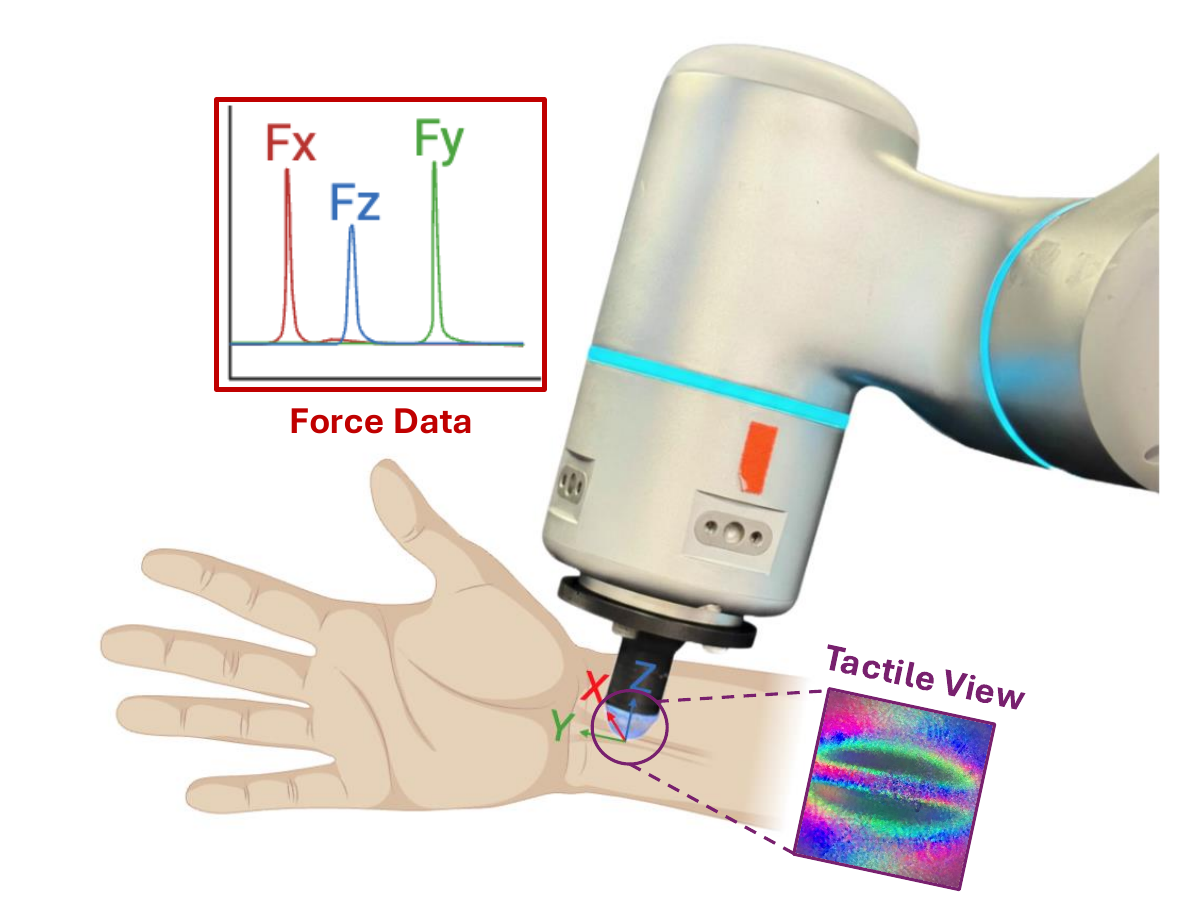}
\caption{PhysioVisionFT (PVFT): a force/torque sensor captures contact force data; a camera captures high-resolution visuotactile images, enabling safe and controlled contact for physiotherapy.
}
\label{fig:concept}\label{fig:fig1}
\vspace{-6mm}
\end{figure}

Vision-based tactile sensors have become popular for providing high-resolution deformation measurements (e.g. \cite{yuan2017gelsight, lambeta2020digit, di2024using, do2022densetact, do2024densetact, lambeta2024digitizing}). However, these are often too compliant or fragile for physiotherapy-scale interactions, where contact forces commonly exceed tens of newtons. Moreover, the estimation of contact forces from deflection patterns can be challenging, especially when in contact with soft tissues that contain subsurface features.

To address these gaps, we present PhysioVisionFT (PVFT), a compact, mechanically robust multimodal sensor that integrates a high-resolution fisheye tactile dome with a 6-axis CoinFT force–torque sensor \cite{choi2025coinft}. This design enables simultaneous acquisition of tactile deformations and dynamic force data under high loading conditions, suitable for subsurface feature detection and physiotherapy.

Using a Flexiv Rizon 4 robot, we performed experiments with prepared tissue phantoms to detect tendon presence, diameter changes, depth variations, crossings, and tendon merging. Across all tasks, tactile imaging revealed clear geometric signatures, while force measurements provided complementary information for stable, and clinically relevant interaction. Our main contributions are:
\vspace{-2mm}
\begin{itemize}\addtolength\itemsep{-2mm}
    \item PVFT, a compact multimodal tool for palpation and physiotherapy, capable of high-force deformation and contact force sensing.
    \item Phantom tissue experiments in which we demonstrate that tactile imaging resolves subsurface geometry that force alone cannot detect, while force/torque sensing provides information for physical interaction.
\end{itemize} 

Together, these results establish the utility of multimodal sensing for robotic palpation and physiotherapy, and lay the groundwork for future medical training, diagnostics, and autonomous therapeutic manipulation.

\section{Sensor Fabrication and Characterization}

\begin{figure}
\centering\includegraphics[width=0.9\linewidth]{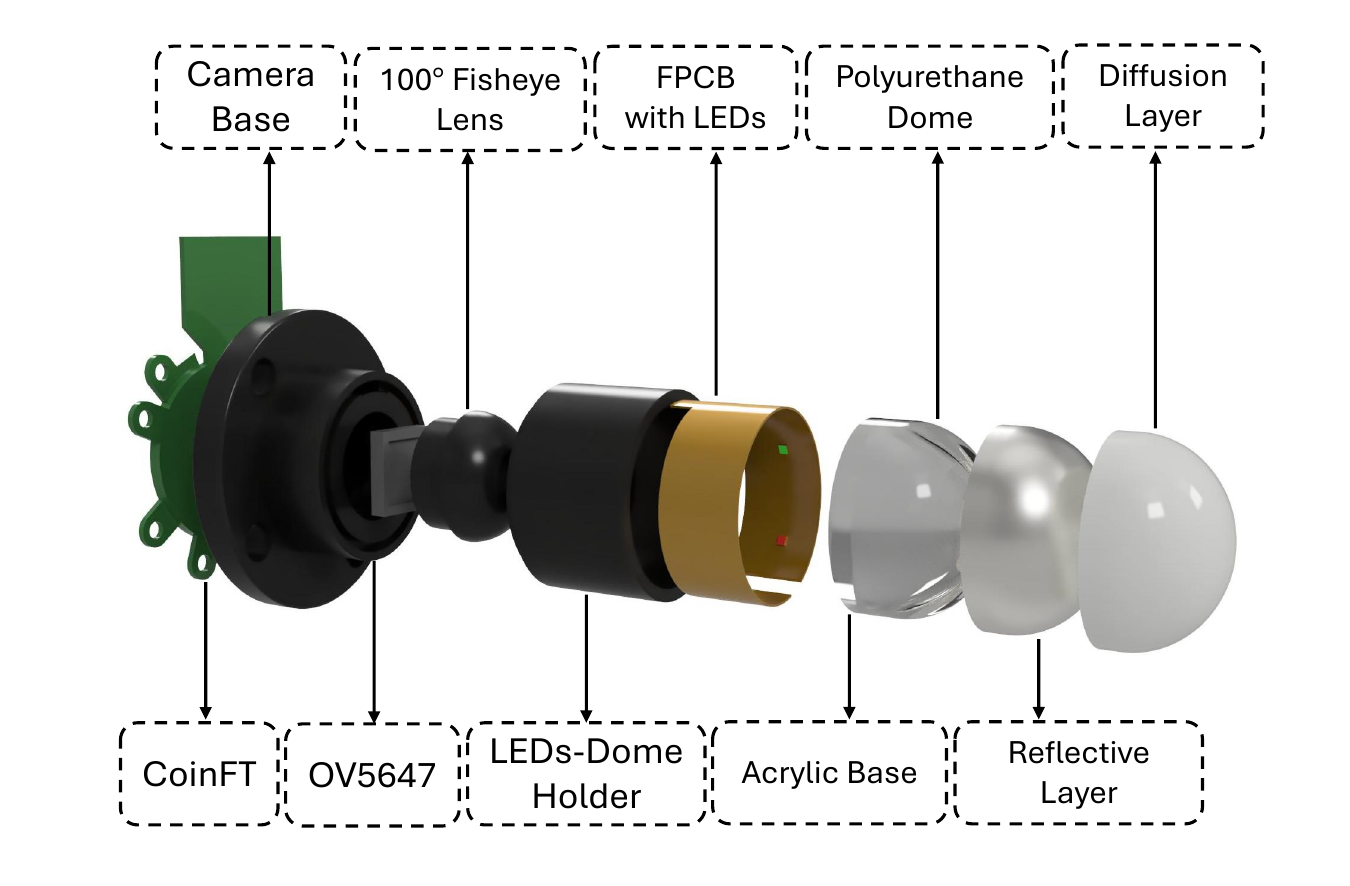}
\caption{Exploded View of the PVFT multimodal sensing tool}
\label{fig:PhysioVisionFT}
\vspace{-4mm}

\end{figure}

The PVFT sensor is constructed through a multi-stage fabrication process that integrates optical, mechanical, and force-sensing components (Fig.~\ref{fig:PhysioVisionFT}). The assembly consists of a fisheye imaging unit, an LED-illuminated dome, and a custom-designed 6-axis force–torque sensor.

\subsection{Fabrication}
A commercial camera (Arducam OV5647) is mounted on a camera base to ensure alignment between the imaging sensor and the dome’s optical center. A 100° fisheye lens is installed after removing the original lens, enabling coverage of the inner dome surface and minimizing distortion near the boundaries.

Illumination is provided by six RGB LEDs (NeoPixel Addressable 1515) soldered to a flexible PCB. The LEDs are addressable and deliver high brightness with low heat generation. The PCB is inserted into a holder, printed using Markforged Onyx, a carbon-fiber–reinforced nylon composite, as the camera base. The material choice provides high stiffness, low mass, and excellent dimensional stability, ensuring precise LED positioning and avoiding unwanted vibrations during robot motion. The matte-black surface finish reduces internal reflections and improves image contrast.

The dome assembly is fabricated as a three-layer optical structure designed to remain robust under high contact loads while capturing deformation details. The primary polyurethane dome is cast using Smooth-On Clear\,Flex\,30 in a mold, which provides a relatively stiff, optically clear shell with controlled wall thickness. This stiffness helps the dome reveal changes in contact patterns under large forces.

After curing, the outer surface of the dome is treated with a primer to activate the polyurethane and promote adhesion. A reflective layer is applied using a mirror-effect spray (Rust-Oleum 301494), forming a thin, uniform coating that enhances diffuse reflections and increases sensitivity to local deformation.

To create the diffusion layer, Clear Flex\,30 is mixed with 2.5\,wt\% white silicone pigment and poured over the reflective surface. After leveling and curing, this semi-opaque layer smooths illumination gradients, suppresses LED hotspots, and improves image contrast. Once the dome assembly is fully cured, it is glued onto the LED–dome holder using hot glue.

\subsection{Characterization}

After mounting the CoinFT sensor beneath the optical dome, we calibrate the PVFT system. To establish accurate force references, the assembled sensor is mounted onto a commercial F/T sensor (ATI Industrial Automation, Gamma) which provides ground-truth force and torque measurements under controlled loading conditions. This setup enables calibration of both the optical and mechanical components in a single pass.


In addition to force calibration with the ATI Gamma, we perform task-specific calibration for high-force interactions using a soft phantom. We employ a 5-layer multilayer perceptron (MLP) model trained with an MSE loss. The calibration procedures include finger-poking tests as well as sliding and rolling tests under pressure. In total, we collect 33434 (23404 for training, 6686 for testing, and 3344 for validation) data points, and the rich interaction set covers a diverse range of changes in the force–torque (FT) data. All tests are conducted using a Flexiv Rizon 4 robot, which supports force control through an impedance control scheme.

\section{Phantom Experiments}

\subsection{Phantom preparation}
Phantoms were prepared to simulate soft tissue with embedded tendons. The material is Smooth-On Ecoflex 00-20 silicone. The tendons are printed from PLA with Young's Modulus of 2.58\,GPa, and experience some elastic deformation as one presses on the silicone above them. Prior to each set of tests we treated the upper surface with mineral oil to reduce friction.


\subsection{Test procedure}

\begin{figure}[h]
\centering\includegraphics[width=0.8\linewidth]{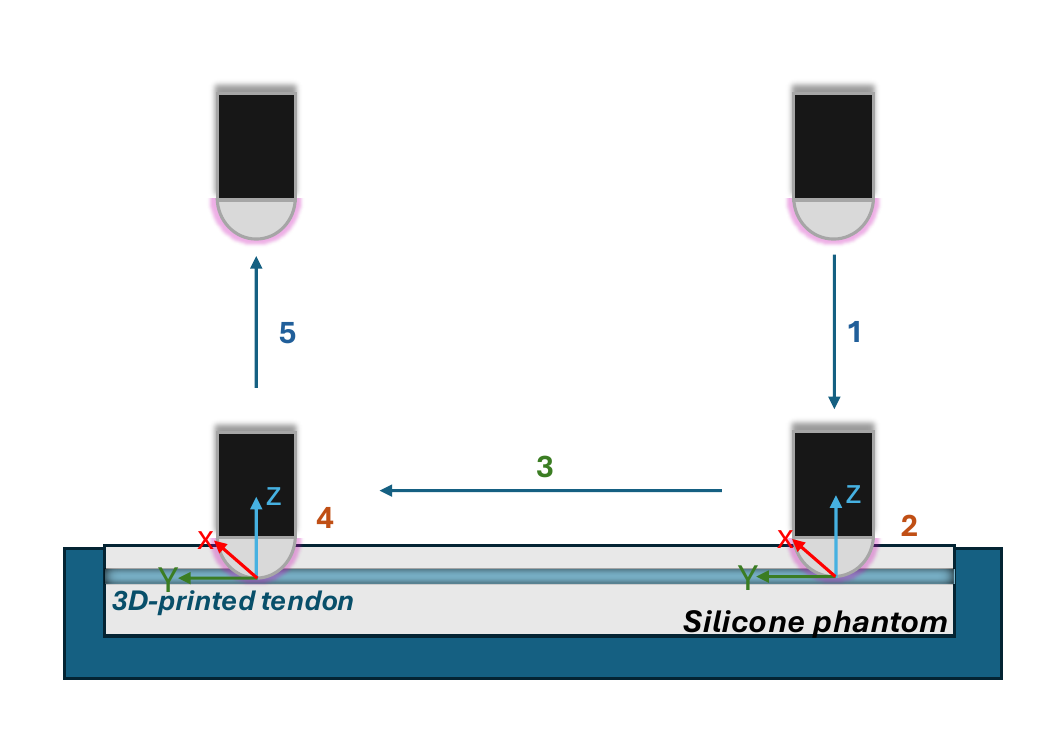}
\vspace{-7mm}
\caption{Procedure: (1) The robot descends along Z direction until the normal force reaches 25\,N, 35\,N or 
45\,N depending on the trial; (2) end-effector holds position for 1\,s; (3) robot translates 120\,mm along the +Y direction under position control; (4) a second 1\,s holding phase; 5) the end-effector retracts and disengages.}
\label{fig: Experimental procedure}
\vspace{-4mm}
\end{figure}

As illustrated in Figure \ref{fig: Experimental procedure}, all experiments follow the same five-step protocol. We use the numeric labels (1–5) here and in subsequent plots to indicate the steps: (1) descent-to-force, (2) and (4) holding intervals, (3) lateral ``ploughing'' motion (sliding while maintaining high force that depresses the phantom tissue), and (5) final retraction.

\begin{figure}[h]
\centering\includegraphics[width=\linewidth]{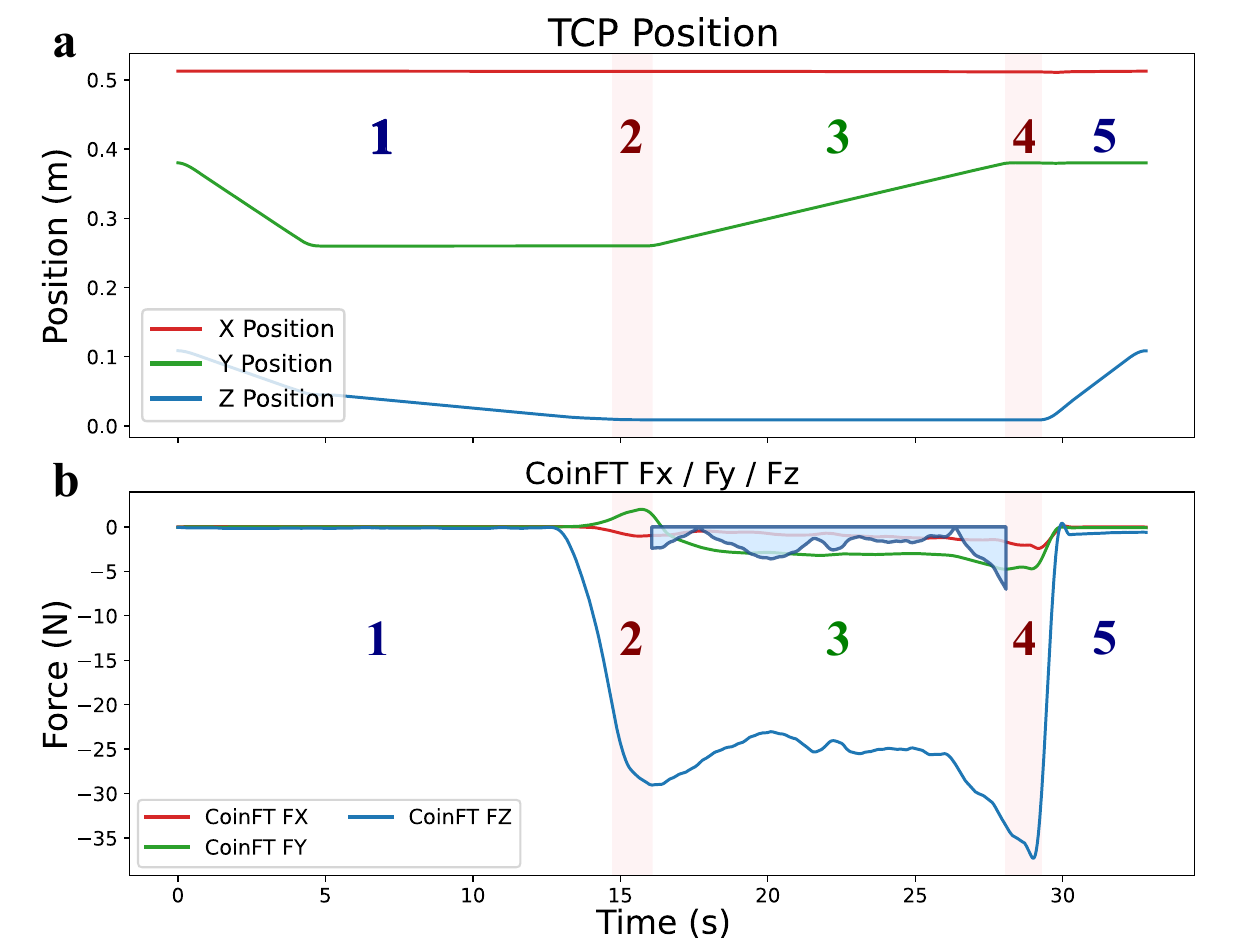}
\caption{Typical motion (a) and force (b) plots corresponding to Fig. \ref{fig: Experimental procedure}, in this case with a single uniform tendon.
The robot descends (1) until it reaches 25\,N normal force. It then dwells (2) for 1\,s and proceeds in the +Y direction (3) at constant Z height. Due to ``ploughing'' of the surface, there is some variation in $F_z$. The robot stops at (4) and departs in (5). Shaded blue region during step (3) shows change in normal force per eq.~\ref{eq:relative force}. 
}
\label{fig:ploughingonUniTendon}
\vspace{-2mm}

\end{figure}

Figure \ref{fig:ploughingonUniTendon} shows an example with steps denoted using the same five labels. Dwell steps (2) and (4) are highlighted with light red shading. Note that between (2) and (4) some variation in force may occur, despite the constant Z position, due to buildup of forces and/or viscoelastic relaxation of the phantom material. 


To quantify how the normal force changes during high-pressure sliding (``ploughing''), we define a force benchmark 
$F_{\text{bench}}$ 
as the mean normal force measured during the first holding phase (region~2). 
The change in normal force is computed as
\begin{equation}
    \Delta F_{z}^{\text{rel}}(t)
    = \frac{F_{z}(t) - F_{\text{bench}}}{F_{\text{bench}}}.
\end{equation}
This metric captures how much the instantaneous normal force deviates from the calibrated reference level. 
In Fig. \ref{fig:ploughingonUniTendon}b and similar plots, the shaded blue region represents the magnitude of force fluctuations during ploughing. We compute the absolute deviation from the benchmark force:
\begin{equation}
    \Delta F_{z}^{\text{abs}}(t) = \left| F_{z}(t) - F_{\text{bench}} \right|.
    \label{eq:relative force}
\end{equation}
For clearer visualization in the plot, we display the negative of this value, $-\Delta F_{z}^{\text{abs}}$.

\subsection{Experiment 1: Transition From Tendon to No-Tendon}
\begin{figure}[h]
\centering\includegraphics[width=\linewidth]{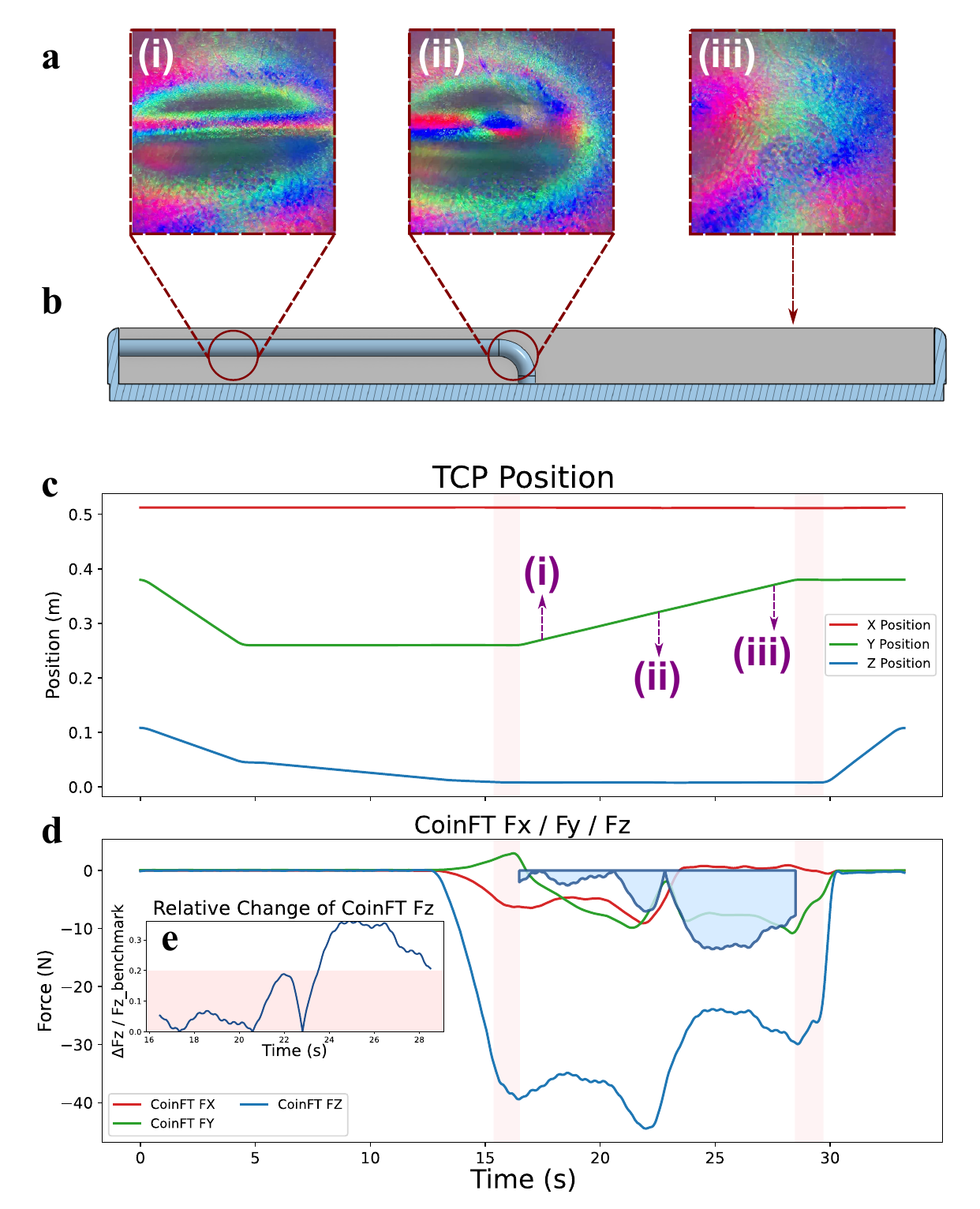}
\caption{Multimodal sensing response when sliding from tendon to no-tendon regions. (a) Shows tactile images corresponding to regions noted in (b). Roman numerals (i), (ii), (ii) show the corresponding times in position (c) and force (d) plots. As in Fig.~\ref{fig:ploughingonUniTendon}, shaded region shows $-\Delta F_{z}^{\text{abs}}$. Inset (e) shows normalized change in $F_z$ during step 3.
}
\label{fig:p1_tendon2notendon25n}
\vspace{-2mm}
\end{figure}

Figure \ref{fig:p1_tendon2notendon25n} shows the system response when transitioning from a tendon into a no-tendon region under a 25\,N initial normal force. The tactile snapshots in (a) exhibit distinct deformation patterns over the tendon compared with the softer silicone region, reflecting the subsurface differences. The paired cross-section schematic in (b) shows where these tactile views were captured.

Panels (c–e) summarize the corresponding robot motion and force measurements. The tool tip trajectory remains consistent, indicating that observed force variations originate from subsurface material differences rather than motion artifacts. As shown in (d), $F_z$ decreases substantially when the tool moves from the tendon region into the softer silicone region, due to reduced surface stiffness.
The relative change plot in (e) quantifies this effect: the deviation from the benchmark force increases significantly when transitioning away from the tendon.

Together, these results confirm that the system can distinguish between stiff (tendon) and soft (no-tendon) regions both in tactile appearance and in force response.

\subsection{Experiment 2: Tendon Height Variation}

\begin{figure}[h]
\centering\includegraphics[width=\linewidth]{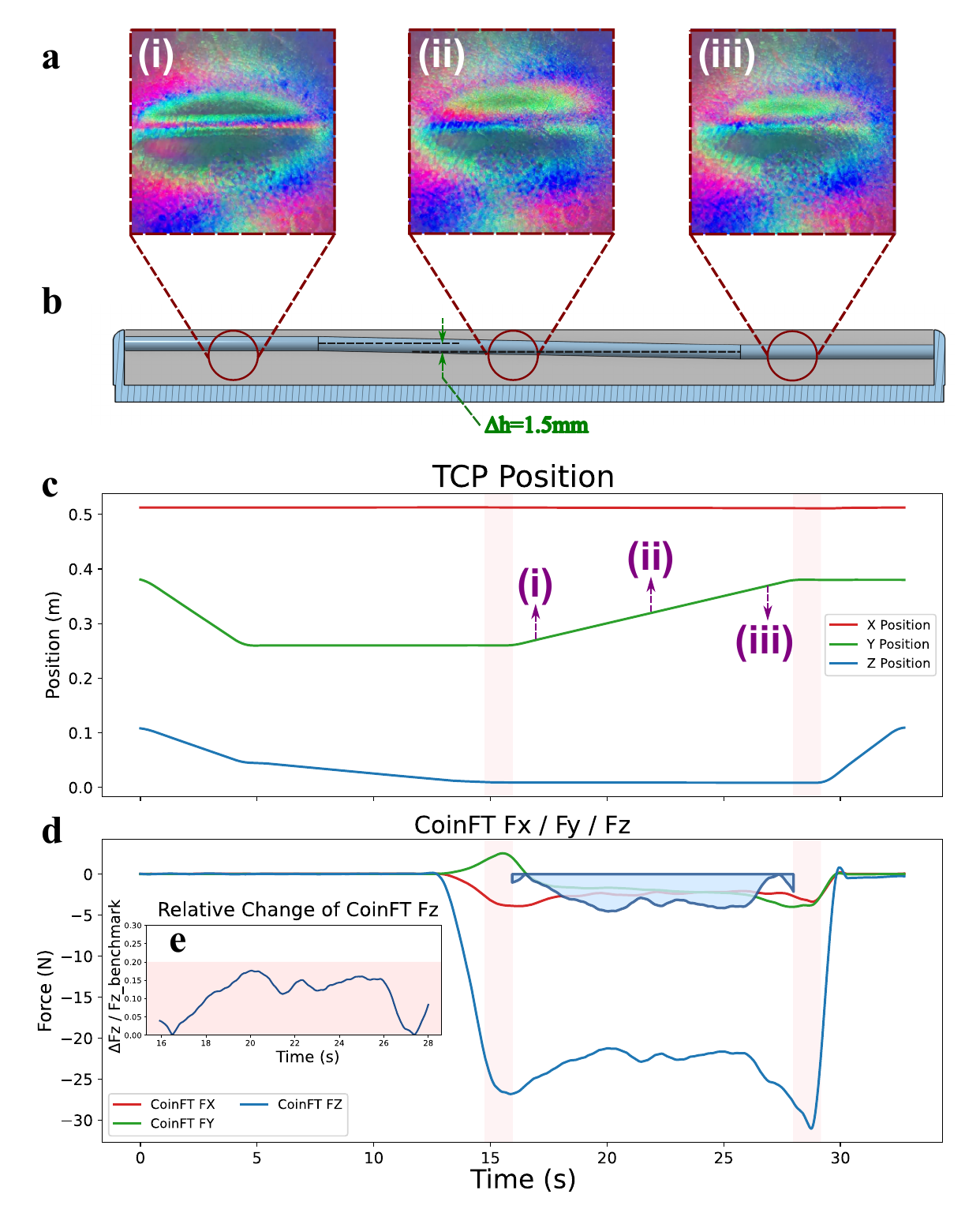}
\caption{Multimodal sensing response to a shallow tendon-height variation ($\Delta$h = 1.5 mm). Labels follow convention of Fig.~\ref{fig:p1_tendon2notendon25n}.}
\label{fig:p2_25}
\vspace{-6.5mm}
\end{figure}

Figure \ref{fig:p2_25} presents an experiment in which the tendon maintains the same width but its centerline height differs by 1.5\,mm between the left and right sides, connected by a sloped transition. Because this variation is shallow, the normal-force response remains relatively unchanged. As shown in (d), the $F_z$ signal exhibits small fluctuations, insufficient to reliably indicate the depth change. In contrast, the tactile views in (a) reveal differences. The region with greater tendon height produces a more pronounced and elongated deformation pattern, while the lower region results in a flatter and more diffused signature. These differences arise from variations in silicone thickness and local dome conformity, making the tactile modality more sensitive than force in this scenario.

\subsection{Experiment 3: Crossing vs. Straight Tendon Geometry}

\begin{figure}[h]
\centering\includegraphics[width=\linewidth]{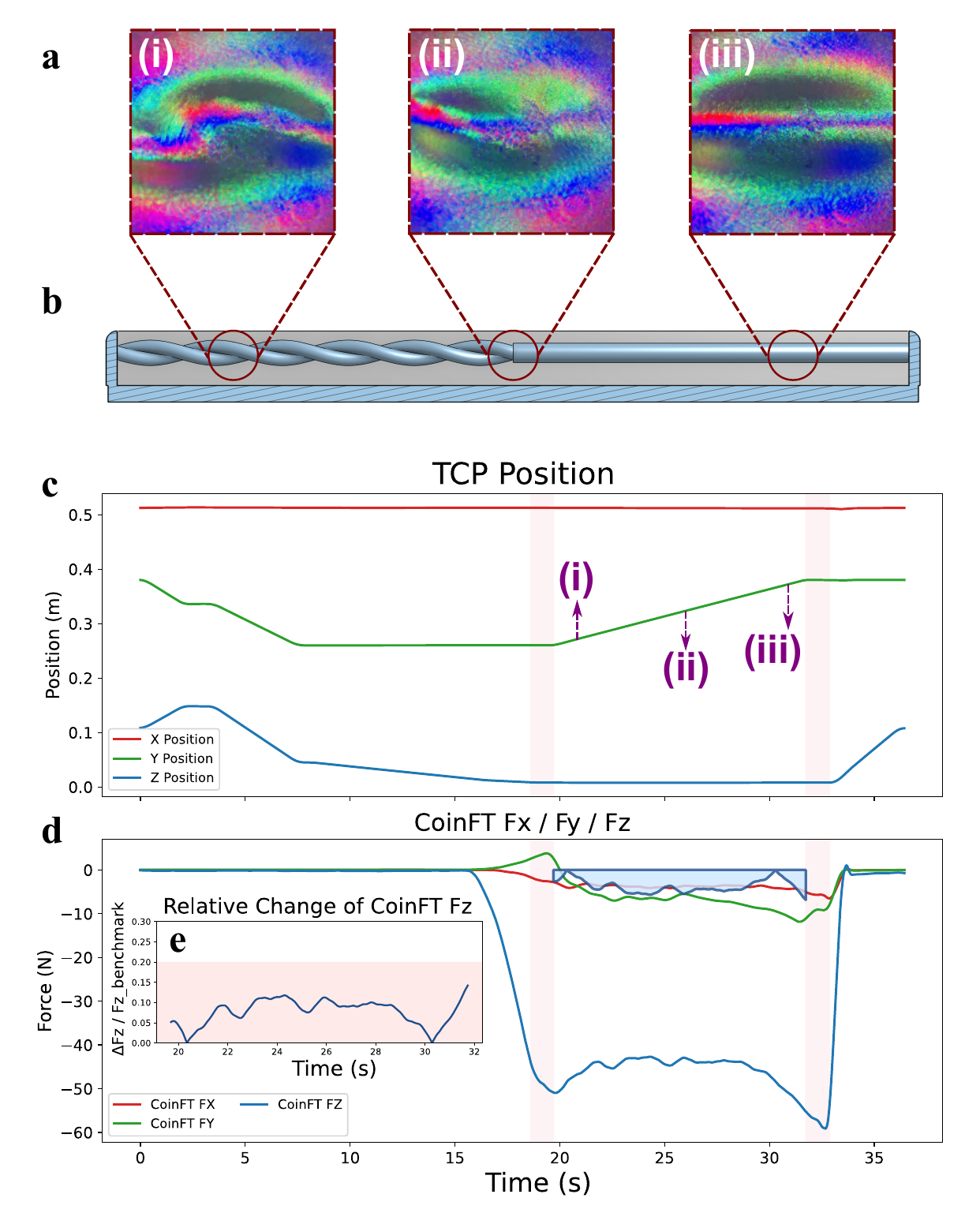}
\caption{Multimodal sensing response when transitioning from a crossed tendon structure to a straight cylindrical tendon.
}
\label{fig:p4_45}
\vspace{-5mm}
\end{figure}

Figure \ref{fig:p4_45} shows an experiment, performed under a higher 45\,N nominal normal force. The tendon geometry begins with a crossed structure and then transitions into a straight 3.5\,mm
cylindrical tendon. Tendon depth is 2.3\,mm below the surface.
 
The tactile views in (a) differentiate the two different regions, showing that details of subsurface geometry are perceivable.
In contrast, the force profile in (d) shows little change across this transition. Similarly, the relative-change curve in (e) shows no distinctive signature.

\subsection{Experiment 4: Distinguishing Tendon Multiplicity}
\begin{figure}[h]
\centering\includegraphics[width=\linewidth]{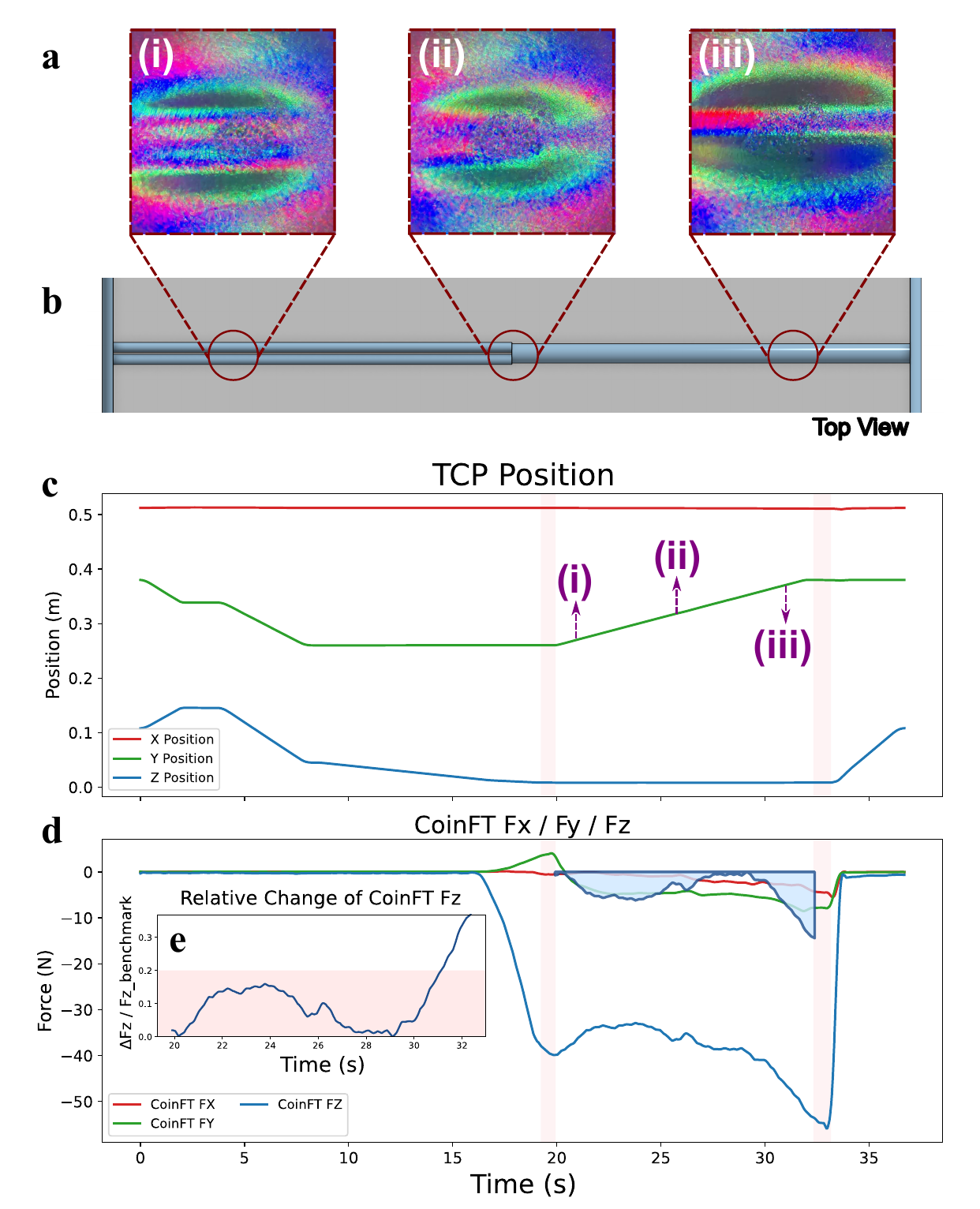}
\caption{Multimodal response when transitioning from two thin tendons to one thick tendon.
Labels follow convention of Fig.~\ref{fig:p1_tendon2notendon25n}.}
\label{fig:p5_35N}
\vspace{-5mm}
\end{figure}

Figure \ref{fig:p5_35N} presents an experiment in which the tendon geometry transitions from two thin parallel tendons into a single thicker tendon, all sharing the same top height, 1.8\,mm below the surface. The top-view geometry shown in (b) highlights the lateral structural differences while keeping the vertical height constant.

The tactile snapshots in (a) clearly distinguish these regions: the two-tendon section produces a split, double-ridge deformation pattern, whereas the single thick tendon generates a broader, more unified footprint. In contrast, the force measurements in (d) show no stable or interpretable pattern across this transition. The $F_Z$ signal fluctuates due to soft-tissue deformation and robot-surface interaction but does not correlate reliably with the number of tendons beneath the sliding dome. The relative change plot in (e) similarly shows variations that are not correlated with the geometry shift. This confirms that force sensing alone is insufficient for detecting lateral structural differences when height and stiffness remain comparable.

Overall, this experiment demonstrates that tactile imaging is essential for identifying how many tendons are present and how they merge, while force signals remain ambiguous.

\subsection{Preliminary Multimodal Training Results}

To evaluate multimodal learning for tendon geometry identification, we collected data using tactile images and force signals. Data collection was performed under impedance control to maintain stable contact with the phantom; additionally, we introduced diverse manual probe movements to increase the variety of interaction patterns. We then trained supervised classification models to compare unimodal baselines against a combined approach using a pretrained ResNet-18 and an attention-based fusion module \cite{lee2019making}. The preliminary performance comparison is summarized in Table \ref{tab:tendon_detection}

We trained supervised classification models to identify tendon geometry (none, single, crossed, or double) from single tactile frames. Data were collected as four dedicated scanning sessions, one per class, in which the probe was rotated from 0 to 180\textdegree{} while maintaining contact with each phantom. After manual cleaning of partial-contact and transition frames, the dataset comprises 7{,}904 labeled frames split 80/10/10 by frame into training, validation, and held-out test sets.

We compare three input modalities: force, tactile image, and combined. For the image path, visual features are extracted from a frozen pretrained ResNet-18 backbone (512-dimensional output). For the force path, the six-axis force/torque vector is processed by a three-layer MLP (64-dimensional output). In the combined model, both feature streams are fused through a cross-modal attention module before a four-class classification head.

All models were trained for 40 epochs with Adam ($\text{lr}=10^{-4}$, cosine annealing). Because the double-tendon class comprises only 6\% of frames, a weighted random sampler ensures balanced class representation in each training batch, and a macro-averaged F1 score is used as the primary evaluation metric. Results on the held-out test set are shown in Table~\ref{tab:tendon_detection}. Force-only classification achieves a macro-F1 of 0.527, confirming the ambiguity of force signals discussed in earlier sections: the model frequently confuses single-tendon and double-tendon regions and misclassifies over half of crossed-tendon frames. Tactile imaging alone raises macro-F1 to 0.835, correctly identifying tendon geometry in nearly all cases but still confuses single and crossed geometries---consistent with their visual similarity in single-frame snapshots. The combined model achieves a macro-F1 of 0.921, substantially reducing confusion between single and crossed tendons and reaching 96.0\% overall accuracy. This indicates that force cues, while insufficient alone, provide complementary information that disambiguates geometrically similar tactile patterns.

\begin{table}[h]
\centering
\caption{Tendon detection performance across input modalities}
\label{tab:tendon_detection}
\begin{tabular}{lccc}
\toprule
\textbf{Modality} & \textbf{Macro-F1} & \textbf{Accuracy}\\
\midrule
Force Only & 0.527  & 61.0\% \\
Tactile Only & 0.835 & 86.9\%\\
Combined & 0.921  & 96.0\% \\
\bottomrule
\end{tabular}
\vspace{-2mm}

\end{table}

Results show that the multimodal model consistently outperforms unimodal baselines. In particular, combining tactile images with force signals improves macro-F1 and overall accuracy compared to image-only models, while force-only models perform substantially worse. Performance gains are most evident for geometrically complex cases, such as crossed tendons, indicating that force sensing provides complementary information that helps disambiguate tactile patterns. These results suggest that multimodal fusion is beneficial in this tendon-geometry classification setting when force cues add information beyond what is available from a single tactile image.
\vspace{-3mm}


\subsection{Force Control for Safe Tissue Interaction}

\begin{figure}
\centering\includegraphics[width=\linewidth]{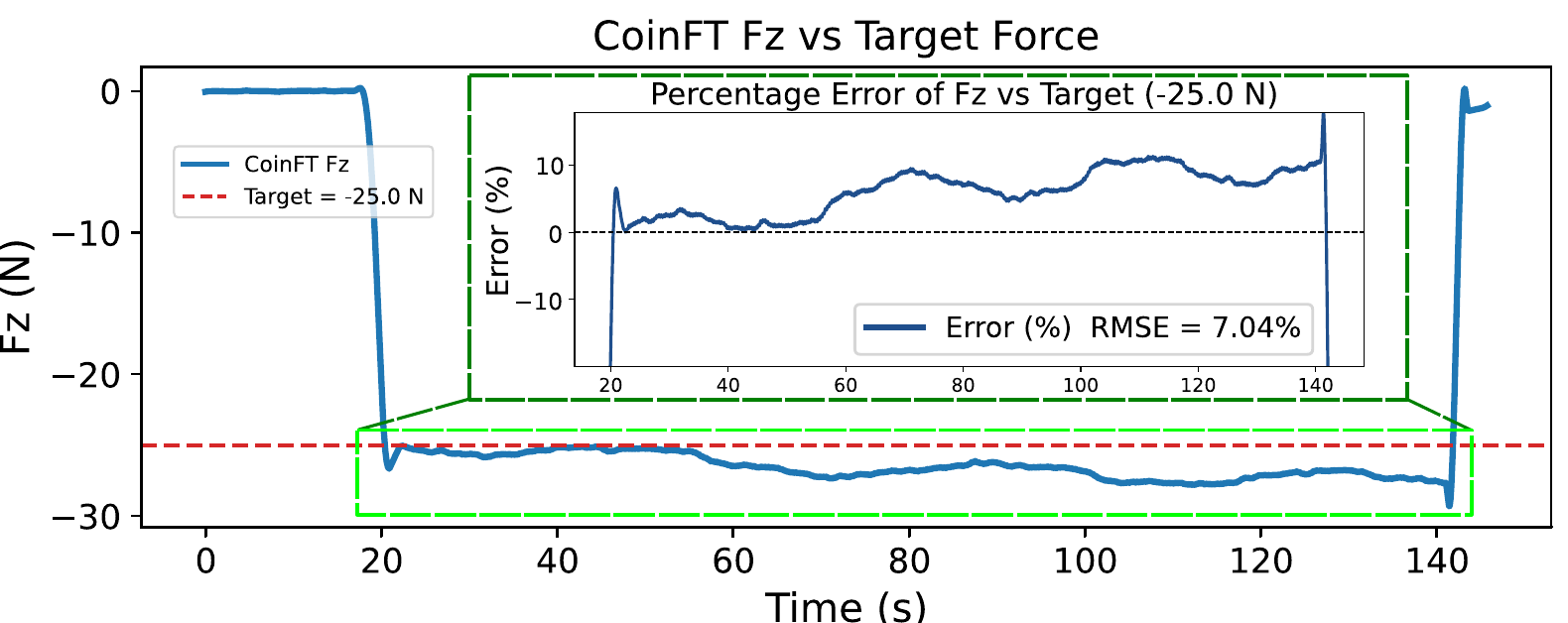}
\caption{CoinFT normal-force tracking performance when maintaining a commanded 25\,N load.
}
\label{fig:massage_application}
\vspace{-5mm}

\end{figure}

Figure \ref{fig:massage_application} shows an experiment for evaluating the role of force tracking from an application perspective. In physical-therapy tasks, maintaining a consistent therapeutic load is important for both treatment quality and patient safety. Here the CoinFT force/torque sensor is particularly useful, as it provides contact force data at 300\,Hz.


We command a –25\,N normal force and use CoinFT to provide data to command a consistent force using an impedance control scheme built into the Flexiv controller. The main plot shows that the measured force remains relatively consistent. However, there is gradual increase in the magnitude of $F_z$ as the ploughing effects of pushing across the soft material start to rotate the tool and introduce relatively large bending moments at the CoinFT sensor, which fall outside the range of calibrated forces and moments. The inset quantifies this effect with a percentage error RMSE of 7.04\%, indicating stable and reliable force tracking.

\section{Conclusion and Future Work}
Our results demonstrate the advantages of combining tactile imaging with force sensing for characterizing subsurface biological structures and supporting physical-therapy interactions. While tactile images provide spatially resolved information about local deformation patterns---revealing tendon presence, diameter changes, depth variations, crossings, and multiplicity---contact force alone often fails to capture these distinctions. In several experiments, even pronounced differences in tendon geometry produced subtle or inconsistent force signatures.

Our observations show that force responses are inherently difficult to predict, even under controlled conditions;
qualitatively different experiment cases produced comparable variations in force. 
In comparison, tactile images revealed clear differences, with qualitatively different images corresponding to different changes in conditions. Nonetheless, we argue that force sensing is valuable as it directly captures therapeutic pressure levels and variations, and provides a basis for force control during sliding and rolling.

\textbf{Future work:} We plan to extend PVFT into a multimodal robot-learning framework that integrates tactile imaging, force feedback, and visual observations from the robot. By collecting demonstrations from expert clinicians and leveraging learning-based policy models such as diffusion policies, the robot could learn closed-loop palpation and treatment strategies that adapt to tendon geometry, depth, and orientation while maintaining safe therapeutic force levels. In parallel, we aim to exploit the tactile images and force signals for learning to infer subsurface structures directly from sensor data, enabling automatic tendon localization and characterization. Together, these directions represent a step toward safer, more repeatable, and more informative robot-assisted palpation and physiotherapy workflows.

\section*{Acknowledgments}

This work originated during Tian-Ao Ren's summer 2025 internship at Symbiokinetics Inc. We thank the Stanford Robotics Center and Symbiokinetics for their collaboration and support throughout this project. We thank Stanley Wang and Venny Kojouharov for their help during the experimental work, and Mingyan Zhang for help with figure preparation.


\bibliographystyle{asmeconf}  
\bibliography{ref}

\end{document}